\documentclass[a4paper,10pt,twocolumn]{article}

\usepackage{fontspec}
\usepackage{geometry}
\usepackage{amsmath,amssymb,bm}
\usepackage{graphicx}
\usepackage{booktabs,array}
\usepackage{caption}
\usepackage{titlesec}
\usepackage{cite}
\usepackage{url}
\usepackage[hidelinks]{hyperref}
\usepackage{microtype}

\hypersetup{
  pdftitle={Vehicle Trajectory Prediction via Neural Fusion of Multiple EKF-Based Trajectory Candidates},
  pdfauthor={Seong-Jun Kim and Seung-Hyun Kong}
}

\renewcommand{\thesection}{\Roman{section}}
\renewcommand{\thesubsection}{\arabic{subsection}}
\titleformat{\section}{\centering\bfseries\fontsize{11}{13}\selectfont}{\thesection.}{0.45em}{}
\titleformat{\subsection}{\bfseries\fontsize{10}{12}\selectfont}{\thesubsection.}{0.45em}{}
\titlespacing*{\section}{0pt}{10pt}{6pt}
\titlespacing*{\subsection}{0pt}{8pt}{4pt}
\renewcommand{\refname}{REFERENCES}
\makeatletter
\renewenvironment{thebibliography}[1]
     {\section*{\refname}%
      \@mkboth{\MakeUppercase\refname}{\MakeUppercase\refname}%
      \fontsize{8}{9.5}\selectfont
      \list{[\arabic{enumiv}]}%
           {\settowidth\labelwidth{[#1]}%
            \leftmargin\labelwidth
            \advance\leftmargin\labelsep
            \setlength{\itemsep}{1pt}%
            \setlength{\parsep}{0pt}%
            \setlength{\topsep}{2pt}%
            \usecounter{enumiv}%
            \let\p@enumiv\@empty
            }%
      \sloppy\clubpenalty4000\@clubpenalty\clubpenalty
      \widowpenalty4000%
      \sfcode`\.\@m}
     {\def\@noitemerr{\@latex@warning{Empty `thebibliography' environment}}%
      \endlist}
\makeatother

\begin{document}
\fontsize{9}{10.8}\selectfont

\twocolumn[{%
\begin{@twocolumnfalse}
\begin{center}
{\fontsize{17}{20}\selectfont\bfseries Vehicle Trajectory Prediction via Neural Fusion of Multiple\\\mbox{EKF-Based Trajectory Candidates}\par}
\vspace{4mm}
{\fontsize{11}{13}\selectfont Seong-Jun Kim$^{1}$ and Seung-Hyun Kong$^{1}$\par}
\vspace{2mm}
{\fontsize{9}{11}\selectfont $^{1}$CCS Graduate School of Mobility, Korea Advanced Institute of Science and Technology (KAIST)\par}
\end{center}
\vspace{2mm}
\noindent\textbf{Abstract:} Predicting the future trajectories of surrounding vehicles in autonomous driving is important for collision risk assessment and safe ego-vehicle path planning. Conventional neural network-based trajectory predictors typically achieve strong prediction performance by exploiting agent history, dynamic scene graphs, and semantic maps. However, in specific motion regimes such as acceleration, deceleration, and turning, these predictors may fail to reflect physically feasible trajectories. To address this issue, this study proposes a framework that fuses the output of Trajectron++, a neural network-based trajectory predictor, with extended Kalman filter (EKF)-based multiple trajectory candidates at a late stage. On the nuScenes dataset, the proposed method reduces the average displacement error and final displacement error of the Trajectron++ robot baseline by 13.7\% and 14.6\%, respectively, without modifying the baseline architecture. These results indicate that EKF-based trajectory candidates can effectively complement neural trajectory prediction through learned fusion.\par
\vspace{2mm}
\vspace{4mm}
\end{@twocolumnfalse}
}]

\section{Introduction}

For an autonomous vehicle to operate safely and smoothly, it must accurately predict the future behavior of surrounding vehicles and pedestrians and incorporate those predictions into path planning and decision making. In interaction-intensive situations such as lane changes, overtaking, and intersection traversal, the future trajectories of nearby vehicles are key inputs for collision-risk assessment and avoidance planning. Vehicle behavior is nevertheless multimodal and uncertain because it depends on driver intent, road geometry, and interactions with surrounding traffic. Trajectory prediction must therefore combine a stable representation of physical vehicle motion with a rich understanding of the driving context.

Vehicle trajectory prediction has developed along two main directions: dynamics-based and neural network-based methods. Dynamics-based methods propagate future states using explicit motion models such as constant-velocity, constant-acceleration, constant-turn-rate, and yaw-rate models. They are computationally simple and interpretable and can generate physically plausible trajectories over short prediction horizons. Prior studies have compared motion models for vehicle tracking and prediction~\cite{r1}, while interacting multiple-model methods and EKFs have been widely used for state estimation under maneuver changes and nonlinear vehicle dynamics~\cite{r3,r4}. However, their simplified motion assumptions make it difficult to capture scene context, including interactions with other agents, road structure, and the future plan of the ego vehicle.

Neural network-based methods instead learn from large-scale driving data and jointly encode agent histories, interactions with surrounding agents~\cite{r17}, map information, and ego-future conditions. Social-LSTM~\cite{r5} and Social-GAN~\cite{r6} predicted multi-agent trajectories through recurrent networks and social-interaction modeling, whereas VectorNet~\cite{r7} strengthened scene representations by vectorizing HD maps and agent dynamics. Trajectron~\cite{r8} and Trajectron++~\cite{r9} subsequently combined dynamic scene graphs, map encoding, ego-future conditioning, and conditional variational autoencoders to model multimodal futures. MultiPath and CoverNet constructed sets of feasible futures using anchor trajectories and trajectory sets, respectively~\cite{r10,r11}; TNT generated candidate trajectories around target points~\cite{r12}; LaneGCN incorporated road structure through lane-graph representations~\cite{r13}; and AgentFormer and MTR used transformer architectures to model spatiotemporal agent interactions and intent~\cite{r14,r15}. Related learning-based approaches have also expanded beyond trajectory prediction to reinforcement learning for driving decisions~\cite{r16}, traffic-signal control~\cite{r20}, deep learning for GNSS enhancement and road-crossing detection~\cite{r18}, and fast acquisition of BOC signals~\cite{r2}. Despite their ability to represent context, neural predictors depend on the training distribution and may fail to select physically plausible alternatives in motion regimes such as rapid acceleration, hard braking, turning, and curve entry.

\begin{figure*}[t]
\centering
\includegraphics[width=0.86\textwidth]{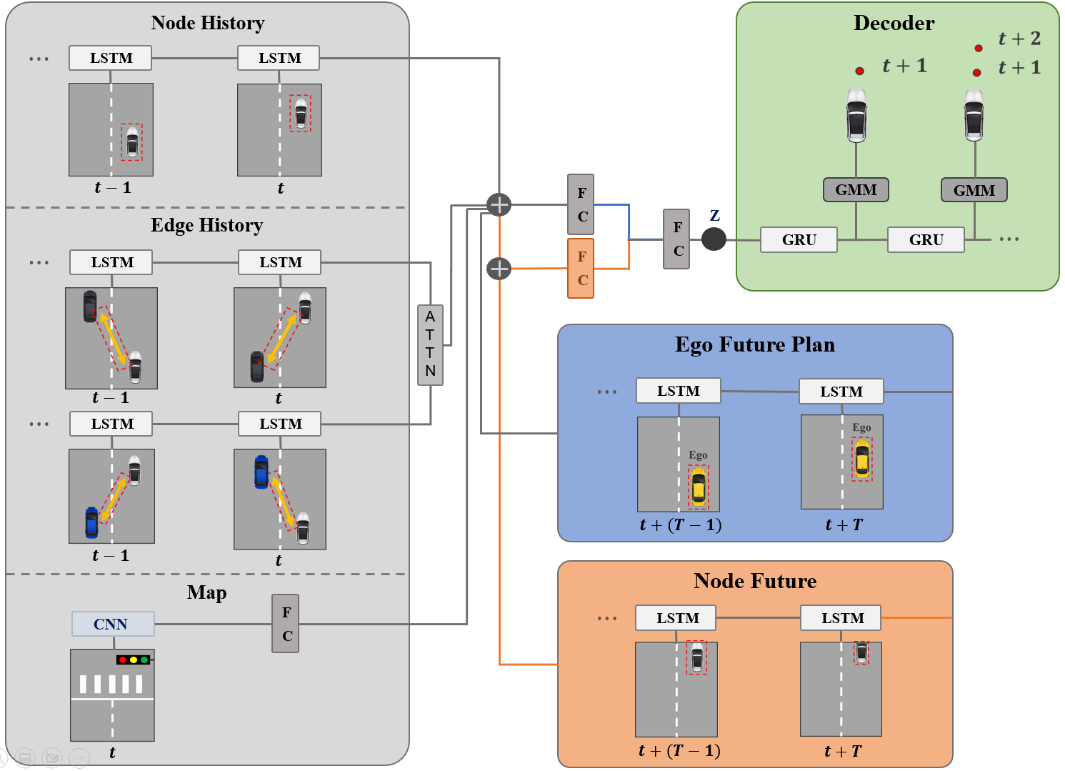}
\caption{Input and prediction architecture of Trajectron++.}
\label{fig:trajectron}
\end{figure*}

The benefits of combining physics-based models with deep neural networks have been demonstrated in autonomous-driving applications~\cite{r19}. Motivated by the complementary strengths of these two approaches, we fuse the output of a learning-based predictor with multiple EKF-based trajectory candidates through a learned late-stage module. This preserves the rich contextual representation of the neural predictor while selectively exploiting physically plausible alternatives in acceleration, deceleration, and turning regimes.

The remainder of this paper is organized as follows. Section~II describes the Trajectron++ baseline. Section~III presents the EKF-based multi-mode candidate generator and neural fusion method. Section~IV reports the experimental setup and evaluation on nuScenes, and Section~V concludes the paper and discusses future work.

\section{Trajectron++ Trajectory Prediction Model}

We use the robot-conditioned configuration of Trajectron++~\cite{r9}, an interaction-aware trajectory prediction model, as the baseline predictor. Figure~\ref{fig:trajectron} summarizes its inputs and prediction flow. The robot configuration predicts the future behavior of surrounding agents conditioned on the planned future trajectory of the ego vehicle.

Trajectron++ represents a scene as a dynamic spatiotemporal graph of nodes and edges. It integrates the target vehicle's history, interactions with surrounding agents, a semantic road map, and the future ego plan to build the contextual features used for prediction. Consequently, the model can account for how neighboring vehicles may react to the future motion of the ego vehicle. Because Trajectron++ captures this rich scene context, it is a suitable baseline for evaluating when and how the proposed EKF-based candidates complement a learning-based predictor.

Each input is processed by a dedicated encoder. Recurrent neural networks encode the target history and interaction histories, a convolutional network encodes the semantic map, and another recurrent network encodes the future ego plan. The resulting representations are combined into a single context vector. Together with the latent variable of a conditional variational autoencoder (CVAE), this context is passed to a GRU-based decoder. Multiple latent samples represent alternative futures, and the decoder sequentially produces a Gaussian-mixture output distribution at each future time step.

We use the Trajectron++ inference result as the baseline candidate in the proposed framework. Among the trajectories produced by the model, the one assigned the highest probability is selected as the representative prediction and denoted by $\mathbf{C}_{0}$. The Trajectron++ weights remain frozen so that any performance change can be attributed to candidate generation and late-stage fusion rather than to modifications of the baseline architecture.

\section{Proposed Method}
\subsection{Overall Architecture}

\begin{figure*}[t]
\centering
\includegraphics[width=0.96\textwidth]{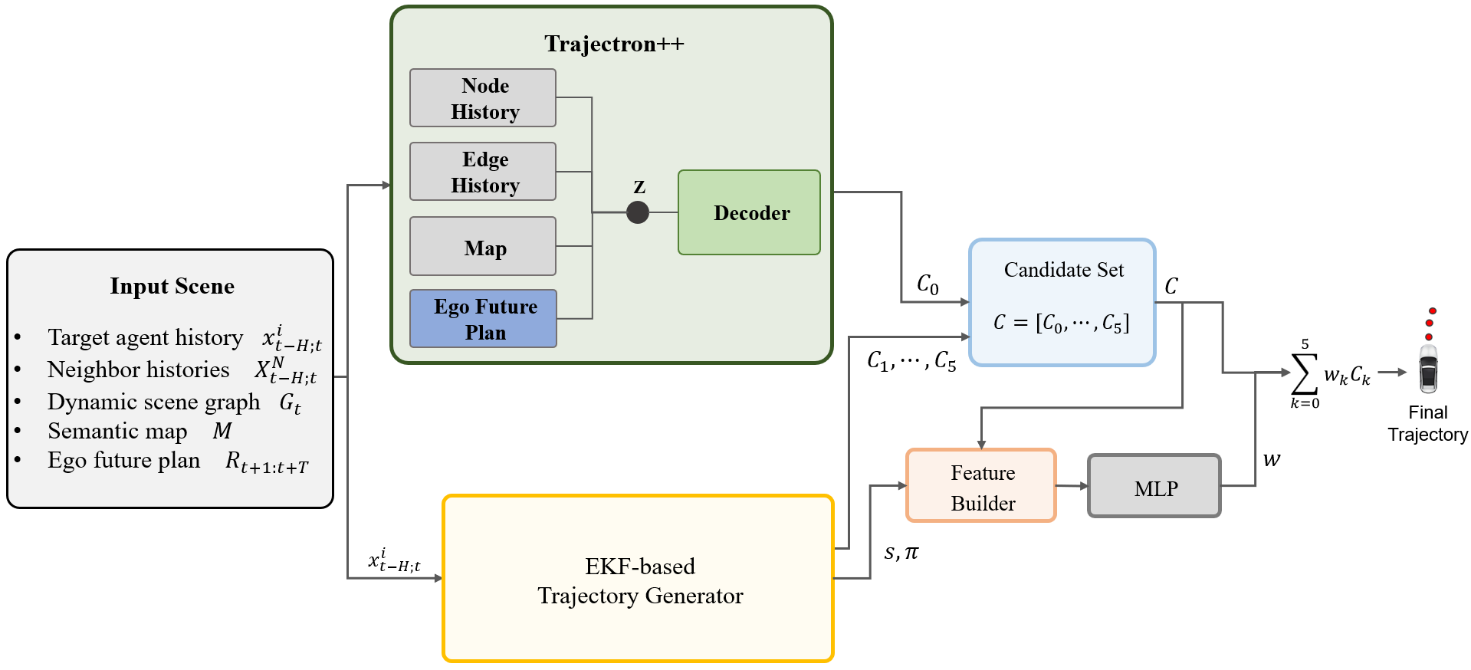}
\caption{Proposed EKF-based multi-mode candidate fusion framework.}
\label{fig:framework}
\end{figure*}

The proposed pipeline, shown in Fig.~\ref{fig:framework}, contains a learning-based prediction branch and a physics-based candidate branch that operate in parallel and are fused at a late stage.

The first branch receives the full scene input---the target and neighbor histories, dynamic scene graph, semantic map, and future ego plan---and uses the Trajectron++ robot model to generate the context-aware neural prediction $\mathbf{C}_{0}$. This is the highest-probability trajectory among the multiple predictions produced by Trajectron++. The second branch uses only the target vehicle's history. Its EKF-based multi-mode generator produces five candidates, $\mathbf{C}_{1},\ldots,\mathbf{C}_{5}$, corresponding to constant velocity, acceleration, deceleration, left turn, and right turn.

The combined set $\mathbf{C}=[\mathbf{C}_{0},\mathbf{C}_{1},\ldots,\mathbf{C}_{5}]$ is processed by a feature builder and a multilayer perceptron (MLP) that predicts sample-specific fusion weights $\mathbf{w}$. The final trajectory is the convex combination $\sum_k w_k\mathbf{C}_k$. Because Trajectron++ remains frozen and only the fusion stage is trained, the method retains the original baseline output while selectively correcting failures in particular motion regimes without retraining the expensive predictor.

\subsection{EKF-Based Multi-Mode Candidate Generation}

The EKF-based generator produces future candidates for several motion modes using only the target vehicle's past trajectory, without surrounding-agent interactions or map context. We use an EKF both to estimate the current state under a nonlinear unicycle model and to initialize candidate rollouts. The EKF handles vehicle motion with heading and yaw rate at low computational cost, is simpler to implement than an unscented Kalman filter or particle filter, and preserves a clear physical interpretation of states and controls. Its limitations are that a single unicycle approximation does not explicitly model steering geometry and that the candidate distribution depends on covariance settings and the mode-scoring design.

\begin{figure*}[t]
\centering
\includegraphics[width=0.98\textwidth]{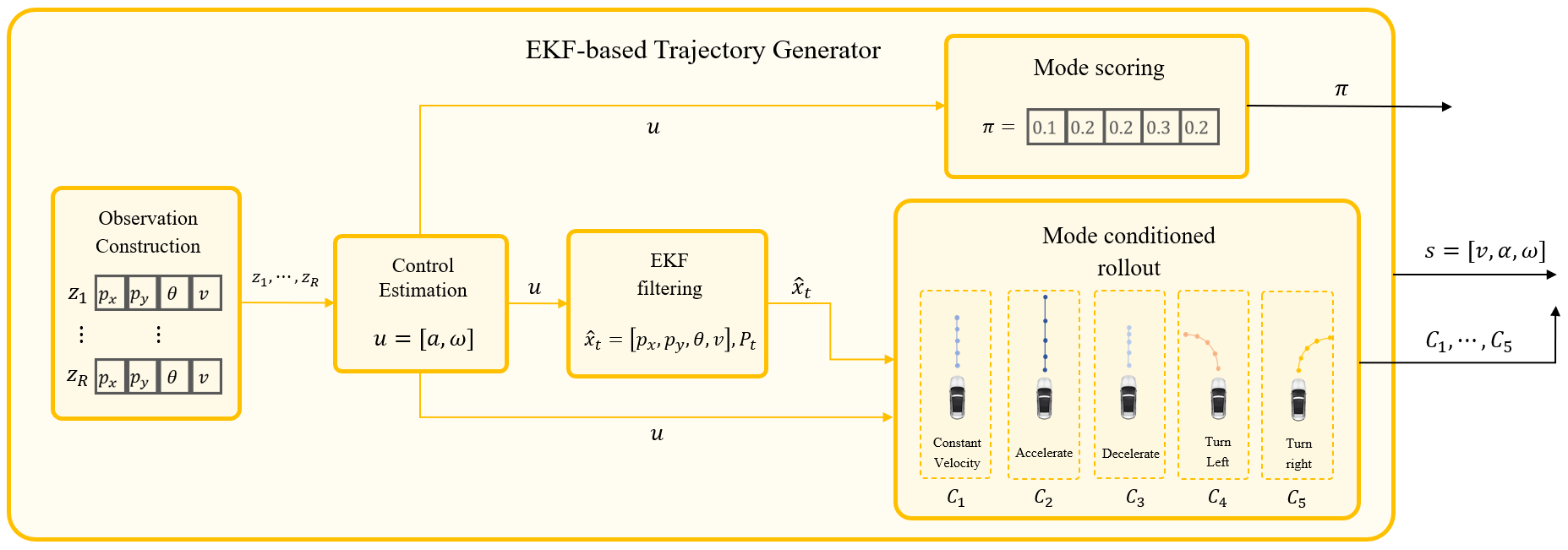}
\caption{Processing pipeline of the EKF-based multi-mode candidate generator.}
\label{fig:generator}
\end{figure*}

The module uses the current state estimated by a single EKF as a common initial condition and rolls out, in parallel, the control input associated with each motion mode. As illustrated in Fig.~\ref{fig:generator}, observations and controls are first constructed from the target history. The EKF prediction and correction steps estimate the current motion state, from which candidates for constant velocity, acceleration, deceleration, left turn, and right turn are generated. The candidate trajectories, mode probabilities, and motion-summary features are then passed to the fusion module.

The target motion is approximated by a unicycle kinematic model that captures longitudinal acceleration, deceleration, and turning over a short horizon. The state vector and control input are

\begin{equation}
\mathbf{x}_{t} = [p_x,\ p_y,\ \theta,\ v]^{\top},
\end{equation}
\begin{equation}
\mathbf{u}_{t} = [a,\ \omega]^{\top},
\end{equation}
where $v$ is scalar speed along the heading direction, $a$ is longitudinal acceleration, and $\omega$ is yaw rate. Because $a$ and $\omega$ must be inferred from the trajectory history without additional sensors, they are computed by finite differences between adjacent time steps:

\begin{equation}
a_{t} = (v_{t}-v_{t-1})/\Delta t,
\end{equation}
\begin{equation}
\omega_{t} = \operatorname{wrap}(\theta_{t}-\theta_{t-1})/\Delta t.
\end{equation}
Here, $\operatorname{wrap}(\cdot)$ normalizes an angular difference to $[-\pi,\pi]$. To suppress unrealistic trajectories caused by observation noise or discretization error, estimated speed, acceleration, and yaw rate are clipped to predefined physical ranges.

The discrete unicycle transition model uses the midpoint heading $\theta_{\mathrm{mid}}$ and midpoint speed $v_{\mathrm{mid}}$ so that heading and speed can change within a time step:

\begin{equation}
p_{x,t+1}=p_{x,t}+v_{\mathrm{mid}}\cos\theta_{\mathrm{mid}}\,\Delta t,
\end{equation}
\begin{equation}
p_{y,t+1}=p_{y,t}+v_{\mathrm{mid}}\sin\theta_{\mathrm{mid}}\,\Delta t,
\end{equation}
\begin{equation}
\theta_{t+1}=\operatorname{wrap}(\theta_t+\omega_t\Delta t),
\end{equation}
\begin{equation}
v_{t+1}=\operatorname{clip}(v_t+a_t\Delta t,0,v_{\max}).
\end{equation}
The observation has the same dimension as the state:

\begin{equation}
\mathbf{z}_{t}=[p_x,\ p_y,\ \theta,\ v]^{\top}.
\end{equation}

The EKF uses this transition model and the observations from the trajectory history to estimate the final state $\widehat{\mathbf{x}}_t$ and covariance $\mathbf{P}_t$. Since no measurements are available over the future horizon, $\widehat{\mathbf{x}}_t$ becomes the initial state for mode-conditioned rollouts.

The observation matrix is set to $\mathbf{H}=\mathbf{I}_4$. The initial, process-noise, and measurement-noise covariances are fixed to $\mathbf{P}_0=\operatorname{diag}(10,10,\pi,5)$, $\mathbf{Q}=\operatorname{diag}(0.25,0.25,0.0225,1.0)$, and $\mathbf{R}=\operatorname{diag}(1.0,1.0,0.0625,2.25)$, respectively. These values account for the different units of position, heading, and speed. No synthetic noise is added to the input history; noise is represented only through covariance propagation and measurement updates in the EKF.

Five motion modes are defined: constant velocity (CV), acceleration (Accel), deceleration (Decel), left turn (Left), and right turn (Right). Let $a$ and $\omega$ be representative controls estimated from recent history. The mode controls are

\begin{align}
\mathrm{CV}:&\quad a=0,\ \omega=0, \\
\mathrm{Accel}:&\quad a=\max(a,1.5),\ \omega=0, \\
\mathrm{Decel}:&\quad a=\min(a,-2.0),\ \omega=0, \\
\mathrm{Left}:&\quad a=0,\ \omega=\max(\omega,0.18), \\
\mathrm{Right}:&\quad a=0,\ \omega=\min(\omega,-0.18).
\end{align}
The CV mode retains current speed and heading. Accel and Decel model longitudinal speed changes, while Left and Right apply positive or negative yaw rates to represent curved motion. Each control is propagated for the prediction horizon $T$ using the transition model above, yielding

\begin{equation}
\mathbf{C}_{k}=\operatorname{rollout}(f,\widehat{\mathbf{x}}_t,\mathbf{u}_k,T).
\end{equation}
Here, $\operatorname{rollout}(\cdot)$ repeatedly applies the transition function $f$. All candidates start from the same estimated state but branch under different control hypotheses, adding physically motivated acceleration, deceleration, and turning alternatives that the neural baseline may miss.

We also compute a mode probability $\pi_k$ that indicates how well each hypothesis agrees with recent motion. Using the magnitude and sign of the representative $a$ and $\omega$, the evidence scores are

\begin{align}
\operatorname{score}_{\mathrm{CV}} &= 1.2-|a|/1.0-|\omega|/0.12,\\
\operatorname{score}_{\mathrm{Accel}} &= (a-a_{\mathrm{thr}})/0.9-|\omega|/0.18,\\
\operatorname{score}_{\mathrm{Decel}} &= (-a-a_{\mathrm{thr}})/0.9-|\omega|/0.18,\\
\operatorname{score}_{\mathrm{Left}} &= (\omega-\omega_{\mathrm{thr}})/0.08-|a|/3.0,\\
\operatorname{score}_{\mathrm{Right}} &= (-\omega-\omega_{\mathrm{thr}})/0.08-|a|/3.0.
\end{align}
The mode distribution is

\begin{equation}
\pi_k=\operatorname{softmax}(\operatorname{score}_k).
\end{equation}
These probabilities do not directly select a candidate or determine the final prediction. They are auxiliary features for the fusion module. The generator ultimately outputs five candidate trajectories $\mathbf{C}_1,\ldots,\mathbf{C}_5$, the mode distribution $\bm{\pi}$, and a motion-summary feature containing speed, acceleration, and yaw rate.

\subsection{Learned Fusion}

The learned fusion module receives the Trajectron++ candidate $\mathbf{C}_0$, the five EKF candidates $\mathbf{C}_1,\ldots,\mathbf{C}_5$, mode probabilities $\bm{\pi}$, and a motion summary $\mathbf{s}$. We first compute the probability-weighted EKF mean

\begin{equation}
\overline{\mathbf{C}}=\sum_{k=1}^{5}\pi_k\mathbf{C}_k.
\end{equation}
The fusion feature $\bm{\varphi}$ concatenates the relative trajectory of each EKF candidate with respect to the baseline, the difference between $\overline{\mathbf{C}}$ and the baseline, the mode probabilities, a one-hot encoding of the most likely mode, and scalar motion-summary features for speed, acceleration, and yaw rate. With prediction horizon $T=6$ and $S=5$ EKF modes, $\bm{\varphi}$ has 99 dimensions: 12 for the baseline, 60 for the five relative trajectories, 12 for the weighted-mean difference, 5 for mode probabilities, 5 for the one-hot mode, and 5 scalar motion features.

The MLP predicts logits for all six candidates and converts them to weights by

\begin{equation}
\mathbf{w}=\operatorname{softmax}(\operatorname{MLP}(\bm{\varphi})).
\end{equation}
The final prediction is

\begin{equation}
\widehat{\mathbf{y}}_t=\sum_{k=0}^{5}w_k\mathbf{C}_{k,t}.
\end{equation}
This convex combination lies within the convex hull of candidate positions at each time step, limiting extreme extrapolation. Dynamic plausibility is encouraged indirectly by the plausibility of the candidates and by the learned fusion weights.

\subsection{Training Objective}

Training minimizes a weighted sum of four losses. The main prediction loss is a SmoothL1 distance between the fused output and ground truth. The candidate with the lowest average displacement error for each sample defines an oracle label, which supervises a cross-entropy loss so that the fusion network learns which candidate to emphasize. A baseline-retention term discourages the baseline weight from collapsing, and a guard loss penalizes samples for which fusion is substantially worse than the baseline. Let $f_i$ and $b_i$ denote the fused and baseline SmoothL1 errors of sample $i$, respectively, and let $m$ be the allowed margin. The losses are

\begin{align}
L_{\mathrm{pred}} &= \operatorname{SmoothL1}(\widehat{\mathbf{y}},\mathbf{Y}),\\
k^* &= \operatorname*{arg\,min}_{k}\operatorname{ADE}(\mathbf{C}_k,\mathbf{Y}),\\
L_{\mathrm{CE}} &= \operatorname{CE}(\bm{\ell},k^*),\\
L_{\mathrm{base}} &= \frac{1}{B}\sum_i(1-w_{i,0}),\\
L_{\mathrm{guard}} &= \frac{1}{B}\sum_i\max(0,f_i-b_i-m),\\
L &= L_{\mathrm{pred}}+\lambda_{\mathrm{CE}}L_{\mathrm{CE}}+\lambda_{\mathrm{base}}L_{\mathrm{base}}+\lambda_{\mathrm{guard}}L_{\mathrm{guard}}.
\end{align}

\section{Experiments and Results}
\subsection{Data and Evaluation Protocol}

Experiments use nuScenes v1.0-trainval with the map expansion pack and evaluate VEHICLE agents in the official validation split~\cite{r21}. Each sample contains 8 history steps and 6 prediction steps, and the evaluation set contains 19,405 samples. Performance is measured by average displacement error (ADE) and final displacement error (FDE):

\begin{equation}
\operatorname{ADE}=\frac{1}{T}\sum_{t=1}^{T}\lVert\widehat{\mathbf{y}}_t-\mathbf{y}_t\rVert_2,
\end{equation}
\begin{equation}
\operatorname{FDE}=\lVert\widehat{\mathbf{y}}_T-\mathbf{y}_T\rVert_2.
\end{equation}

The Trajectron++ baseline uses a frozen epoch-20 checkpoint trained on the nuScenes training split. The fusion MLP is trained using baseline predictions and EKF candidates from 601 training scenes (70,934 samples), and its checkpoint is selected using a separate set of 70 held-out scenes (9,274 samples). The official validation split of 142 scenes (19,405 samples) is used only for final evaluation. Because stable estimates of speed, acceleration, and yaw rate require sufficient history, all methods are evaluated on samples with the same 8-step history. Reported baseline values are reproduced under the same code, data, and hyperparameter settings used for the proposed method.

\subsection{Implementation Details}

The baseline uses the Trajectron++ nuScenes robot configuration, including map encoding and future ego conditioning, with a prediction horizon of 6, maximum history length of 8, batch size of 512, and an epoch-20 checkpoint. All trajectories are represented in a target-centered relative coordinate frame.

\subsection{Quantitative Results}

Table~\ref{tab:main} compares the baseline and proposed method. EKF-based multi-mode fusion reduces ADE from 0.9035~m to 0.7793~m and FDE from 1.8618~m to 1.5910~m, corresponding to relative improvements of 13.74\% and 14.55\%, respectively. These gains are obtained by adding physics-based candidates and learned fusion while preserving the neural baseline output and architecture.

\begin{table}[t]
\centering
\caption{Quantitative comparison of the baseline and proposed method (ADE/FDE in meters).}
\label{tab:main}
\begin{tabular}{lcc}
\toprule
\textbf{Model} & \textbf{ADE} & \textbf{FDE} \\
\midrule
Trajectron++ & 0.9035 & 1.8618 \\
\textbf{Proposed} & \textbf{0.7793} & \textbf{1.5910} \\
\bottomrule
\end{tabular}
\end{table}

\subsection{Applicability to Different Trajectory Predictors}

To evaluate model-independent applicability, we freeze the official checkpoints of CMT~\cite{r22}, LAformer~\cite{r23}, and AgentFormer~\cite{r14} and apply the same EKF candidates and lightweight fusion module. Table~\ref{tab:transfer} reports the reduction in mean error before and after fusion for each predictor.

CMT and LAformer are evaluated using ADE/FDE of their highest-probability single trajectory, whereas AgentFormer uses minimum ADE/FDE over five modes. The table therefore compares within-model improvements rather than absolute error values across models. CMT and AgentFormer are evaluated in three runs using their official train/train-val/validation splits; the LAformer result is an auxiliary scene-disjoint analysis within its validation split.

Fusion reduces CMT top-1 ADE/FDE by $2.00\pm0.06\%$ and $2.07\pm0.06\%$, LAformer top-1 by 1.68\% and 0.46\%, and AgentFormer $K=5$ by $0.45\pm0.01\%$ and $0.62\pm0.01\%$. Thus, the module generalizes across prediction architectures and output formats, although the size of the gain varies by model.

\begin{table}[t]
\centering
\caption{Mean error reduction after fusion across different trajectory prediction models.}
\label{tab:transfer}
\resizebox{\columnwidth}{!}{%
\begin{tabular}{lccc}
\toprule
\textbf{Prediction model} & \textbf{Evaluation set} & \textbf{ADE reduction} & \textbf{FDE reduction} \\
\midrule
CMT top-1 (2025) & nuScenes val (9,041) & $2.00\pm0.06\%$ & $2.07\pm0.06\%$ \\
LAformer top-1 (2024) & Scene-disjoint test (1,947) & 1.68\% & 0.46\% \\
AgentFormer $K=5$ (2021) & nuScenes val (9,041) & $0.45\pm0.01\%$ & $0.62\pm0.01\%$ \\
\bottomrule
\end{tabular}}
\end{table}

\subsection{Oracle Upper Bound and Candidate Analysis}

Table~\ref{tab:oracle} evaluates the performance available if the best candidate is selected for each sample. The mode-only oracle selects among the five EKF candidates, while the baseline-inclusive oracle also considers $\mathbf{C}_0$. The mode-probability weighted EKF average is a nonlearned combination and does not outperform the baseline, whereas the learned MLP fusion does. The best EKF-mode oracle reaches 0.7434~m ADE, and the baseline-inclusive oracle reaches 0.4998~m. This gap indicates that further gains depend more on improved candidate selection than on candidate generation alone.

\begin{table}[t]
\centering
\caption{Oracle (best-candidate) upper-bound analysis (meters).}
\label{tab:oracle}
\resizebox{\columnwidth}{!}{%
\begin{tabular}{lcc}
\toprule
\textbf{Method} & \textbf{ADE} & \textbf{FDE} \\
\midrule
Multi-mode candidate oracle & 0.7434 & 1.4475 \\
Baseline + multi-mode candidate oracle & 0.4998 & 1.0042 \\
\bottomrule
\end{tabular}}
\end{table}

Table~\ref{tab:candidates} reports, for each candidate, how often it is the oracle choice, its mean fusion weight, and its mean ADE. CV and the baseline receive the largest shares. Acceleration, deceleration, and turning candidates have larger average errors but complement the baseline in specific failure cases. Notably, CV has a worse mean ADE (1.1767~m) than the baseline (0.9035~m) but is the best candidate for many individual samples, showing that average candidate quality and sample-wise usefulness are distinct. An EKF candidate outperforms the baseline in 76.78\% of samples, and learned fusion outperforms the baseline in 67.55\% of samples.

\begin{table}[t]
\centering
\caption{Per-candidate oracle selection rate, mean fusion weight, and candidate ADE.}
\label{tab:candidates}
\resizebox{\columnwidth}{!}{%
\begin{tabular}{lccc}
\toprule
\textbf{Candidate} & \textbf{Oracle selection} & \textbf{Mean fusion} & \textbf{ADE} \\
\midrule
Trajectron++ & 23.22\% & 27.42\% & 0.9035 \\
CV & 50.22\% & 49.41\% & 1.1767 \\
Accel & 7.47\% & 5.96\% & 3.3062 \\
Decel & 12.48\% & 11.24\% & 1.9824 \\
Left & 3.27\% & 2.97\% & 1.6862 \\
Right & 3.34\% & 2.99\% & 1.7005 \\
\bottomrule
\end{tabular}}
\end{table}

\subsection{Mode-Probability Analysis}

\begin{figure}[t]
\centering
\includegraphics[width=0.96\columnwidth]{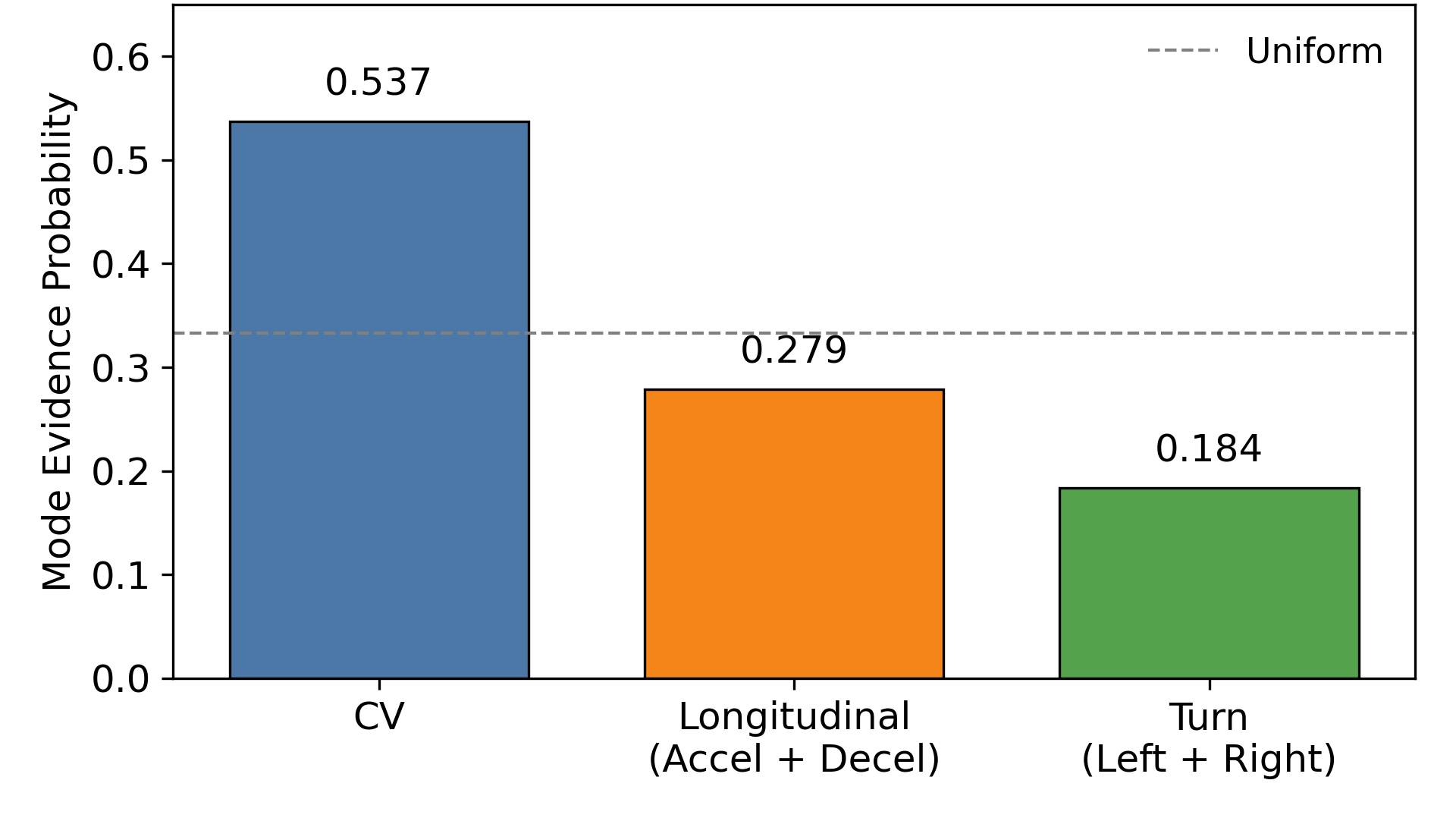}
\caption{Mean mode probability for constant-velocity, longitudinal, and turning regimes.}
\label{fig:modeprob}
\end{figure}

The EKF module estimates the target's recent motion state and produces probabilities for CV, acceleration, deceleration, left turn, and right turn. These probabilities are not final decisions; they serve as auxiliary motion-evidence features for the fusion network. Figure~\ref{fig:modeprob} aggregates them over the validation set into constant-velocity, longitudinal, and turning groups. The dashed line marks a uniform group probability of 0.333.

Because most nuScenes vehicle trajectories are close to constant velocity over the short prediction horizon, the CV group has the highest mean probability, 0.537. The longitudinal and turning groups have mean probabilities of 0.279 and 0.184, respectively, and together account for 0.463 of the probability mass. Thus, the EKF module still supplies substantial non-CV motion evidence. The dominant CV probability reflects the data distribution rather than a limitation of the method; performance gains arise from the candidate set and learned fusion rather than from the mode probabilities alone.

\subsection{Qualitative Results}

\begin{figure*}[t]
\centering
\includegraphics[width=0.96\textwidth]{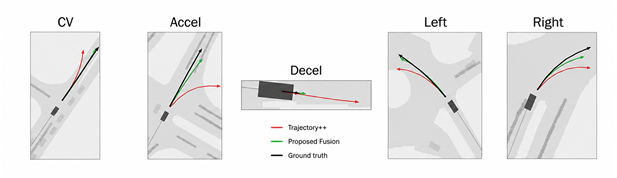}
\caption{Qualitative comparison of trajectory predictions by motion regime.}
\label{fig:qualitative}
\end{figure*}

Figure~\ref{fig:qualitative} shows representative predictions from five motion regimes. Red denotes the Trajectron++ baseline, green the fused output, and black the ground-truth future. Across the examples, the proposed method produces trajectories closer to ground truth. For CV, fusion corrects endpoint and travel-distance error. For acceleration and deceleration, EKF candidates correct baseline over- or under-prediction in the longitudinal direction. For left and right turns, fusion more accurately reproduces curvature and lateral displacement that the baseline misses. The improvements are most pronounced when the baseline's motion regime is mismatched, demonstrating that physics-based candidates can complement the neural output and that learned fusion can select a better trajectory adaptively.

\section{Conclusion and Future Work}

This paper presented a lightweight framework that augments a learning-based trajectory predictor with multiple physics-based EKF candidates. Without substantially modifying the Trajectron++ baseline, the method improves ADE and FDE on nuScenes by 13.7\% and 14.6\%, respectively. Experiments with other trajectory prediction models further demonstrate applicability across architectures, although the magnitude of improvement varies with the predictor and output format. Oracle analysis shows that the learned method approaches the EKF-only candidate upper bound but remains well above the upper bound available when the baseline is also considered. Better candidate selection and fusion therefore remain the primary directions for future work.


\begin{thebibliography}{99}
\bibitem{r1} R. Schubert, E. Richter, and G. Wanielik, ``Comparison and evaluation of advanced motion models for vehicle tracking,'' Proceedings of the International Conference on Information Fusion, pp. 1--6, 2008.
\bibitem{r2} B. Kim and S.-H. Kong, ``Two-dimensional compressed correlator for fast acquisition of BOC($m,n$) signals,'' IEEE Transactions on Vehicular Technology, vol. 63, no. 6, pp. 2662--2672, 2014.
\bibitem{r3} H. A. P. Blom and Y. Bar-Shalom, ``The interacting multiple model algorithm for systems with Markovian switching coefficients,'' IEEE Transactions on Automatic Control, vol. 33, no. 8, pp. 780--783, Aug. 1988.
\bibitem{r4} M. T. Abbas, M. A. Jibran, M. Afaq, and W.-C. Song, ``An adaptive approach to vehicle trajectory prediction using multimodel Kalman filter,'' Transactions on Emerging Telecommunications Technologies, vol. 31, no. 5, e3734, 2020.
\bibitem{r5} A. Alahi, K. Goel, V. Ramanathan, A. Robicquet, L. Fei-Fei, and S. Savarese, ``Social LSTM: Human trajectory prediction in crowded spaces,'' Proceedings of the IEEE Conference on Computer Vision and Pattern Recognition, pp. 961--971, 2016.
\bibitem{r6} A. Gupta, J. Johnson, L. Fei-Fei, S. Savarese, and A. Alahi, ``Social GAN: Socially acceptable trajectories with generative adversarial networks,'' Proceedings of the IEEE Conference on Computer Vision and Pattern Recognition, pp. 2255--2264, 2018.
\bibitem{r7} J. Gao, C. Sun, H. Zhao, et al., ``VectorNet: Encoding HD maps and agent dynamics from vectorized representation,'' Proceedings of the IEEE/CVF Conference on Computer Vision and Pattern Recognition, pp. 11525--11533, 2020.
\bibitem{r8} B. Ivanovic and M. Pavone, ``The Trajectron: Probabilistic multi-agent trajectory modeling with dynamic spatiotemporal graphs,'' Proceedings of the IEEE/CVF International Conference on Computer Vision, pp. 2375--2384, 2019.
\bibitem{r9} T. Salzmann, B. Ivanovic, P. Karasev, and M. Pavone, ``Trajectron++: Dynamically-feasible trajectory forecasting with heterogeneous data,'' Proceedings of the European Conference on Computer Vision, pp. 683--700, 2020.
\bibitem{r10} Y. Chai, B. Sapp, M. Bansal, and D. Anguelov, ``MultiPath: Multiple probabilistic anchor trajectory hypotheses for behavior prediction,'' Proceedings of the Conference on Robot Learning, PMLR, vol. 100, pp. 86--99, 2020.
\bibitem{r11} T. Phan-Minh, E. C. Grigore, F. A. Boulton, O. Beijbom, and E. M. Wolff, ``CoverNet: Multimodal behavior prediction using trajectory sets,'' Proceedings of the IEEE/CVF Conference on Computer Vision and Pattern Recognition, pp. 14074--14083, 2020.
\bibitem{r12} H. Zhao, J. Gao, T. Lan, C. Sun, B. Sapp, B. Varadarajan, Y. Shen, Y. Shen, Y. Chai, C. Schmid, C. Li, and D. Anguelov, ``TNT: Target-driven trajectory prediction,'' Proceedings of the Conference on Robot Learning, PMLR, vol. 155, pp. 895--904, 2021.
\bibitem{r13} M. Liang, B. Yang, R. Hu, Y. Chen, R. Liao, S. Feng, and R. Urtasun, ``Learning lane graph representations for motion forecasting,'' Proceedings of the European Conference on Computer Vision, pp. 541--556, 2020.
\bibitem{r14} Y. Yuan, X. Weng, Y. Ou, and K. Kitani, ``AgentFormer: Agent-aware transformers for socio-temporal multi-agent forecasting,'' Proceedings of the IEEE/CVF International Conference on Computer Vision, pp. 9813--9823, 2021.
\bibitem{r15} S. Shi, L. Jiang, D. Dai, and B. Schiele, ``Motion Transformer with global intention localization and local movement refinement,'' Advances in Neural Information Processing Systems, vol. 35, pp. 6531--6543, 2022.
\bibitem{r16} S.-H. Kong, I. M. A. Nahrendra, and D.-H. Paek, ``Enhanced off-policy reinforcement learning with focused experience replay,'' IEEE Access, vol. 9, pp. 93152--93164, 2021.
\bibitem{r17} K. Kim, G. Mun, G. Lee, D. Kim, and H. Kim, ``Cooperative lane merge system through driving trajectory sharing based on 5G-NR-V2X,'' Journal of Institute of Control, Robotics and Systems, vol. 30, no. 4, pp. 492--499, 2024.
\bibitem{r18} S. J. Cho, B. S. Kim, T. S. Kim, and S.-H. Kong, ``Enhancing GNSS performance and detection of road crossing in urban area using deep learning,'' Proceedings of the IEEE Intelligent Transportation Systems Conference, pp. 2115--2120, 2019.
\bibitem{r19} S. Kim, J. Choi, D. Jeong, and S. Lee, ``LSTM-based Kalman filter error compensation for enhanced moving object tracking,'' Journal of Institute of Control, Robotics and Systems, vol. 31, no. 1, pp. 68--76, 2025.
\bibitem{r20} C. J. Choe, S. Baek, B. Woon, and S.-H. Kong, ``Deep Q learning with LSTM for traffic light control,'' Proceedings of the Asia-Pacific Conference on Communications, pp. 331--336, 2018.
\bibitem{r21} H. Caesar, V. Bankiti, A. H. Lang, et al., ``nuScenes: A multimodal dataset for autonomous driving,'' Proceedings of the IEEE/CVF Conference on Computer Vision and Pattern Recognition, pp. 11621--11631, 2020.
\bibitem{r22} T. Jiang, Q. Dong, Y. Ma, X. Ji, and Y. Liu, ``Customizable multimodal trajectory prediction via nodes of interest selection for autonomous vehicles,'' Expert Systems with Applications, article 128222, 2025.
\bibitem{r23} M. Liu, H. Cheng, L. Chen, H. Broszio, J. Li, R. Zhao, M. Sester, and M. Y. Yang, ``LAformer: Trajectory prediction for autonomous driving with lane-aware scene constraints,'' Proceedings of the IEEE/CVF Conference on Computer Vision and Pattern Recognition Workshops, pp. 2039--2049, 2024.
\end{thebibliography}
\end{document}